\documentclass[option]{llncs}
\usepackage{graphicx}
\usepackage{amsmath,amssymb} 

 \usepackage[width=122mm,left=12mm,paperwidth=146mm,height=193mm,top=12mm,paperheight=217mm]{geometry}
\usepackage{color}

\usepackage{url}
\usepackage{hyperref}

\usepackage{booktabs}
\usepackage{subfigure}
\begin{document}
\mainmatter

\title{Tracking Small and Fast Moving Objects: A Benchmark} 

\author{\small Zhewen Zhang$^{\dagger}$, Fuliang Wu$^{\dagger}$, Yuming Qiu, Jingdong Liang, Shuiwang Li\thanks{Corresponding Author. $^{\dagger}$ These authors contributed equally. \\E-mail address: zwzhang0101@163.com (Zhewen Zhang), wufuliang@glut.edu.cn (Fuliang Wu),  qiuyuming0706@163.com (Yuming Qiu), liangjingdong@glut.edu.cn (Jingdong Liang), and lishuiwang0721@163.com (Shuiwang Li)}}
\institute{Guilin University of Technology, China}

\maketitle

\begin{abstract}
With more and more large-scale datasets available for training, visual tracking has made great progress in recent years. However, current research in the field mainly focuses on tracking generic objects. In this paper, we present TSFMO, a benchmark for \textbf{T}racking \textbf{S}mall and \textbf{F}ast \textbf{M}oving \textbf{O}bjects. This benchmark aims to encourage research in developing novel and accurate methods for this challenging task particularly. TSFMO consists of 250 sequences with about 50k frames in total. Each frame in these sequences is carefully and manually annotated with a bounding box. To the best of our knowledge, TSFMO is the first benchmark dedicated to tracking small and fast moving objects, especially connected to sports. To understand how existing methods perform and to provide comparison for future research on TSFMO, we extensively evaluate 20 state-of-the-art trackers on the benchmark. The evaluation results exhibit that more effort are required to improve tracking small and fast moving objects. Moreover, to encourage future research, we proposed a novel tracker S-KeepTrack which surpasses all 20 evaluated approaches. By releasing TSFMO, we expect to facilitate future researches and applications of tracking small and fast moving objects. The TSFMO and evaluation results as well as S-KeepTrack are available at \url{https://github.com/CodeOfGithub/S-KeepTrack}.

\end{abstract}

\section{Introduction}
Object tracking is one of the most fundamental problems in computer vision with a variety of applications, including video surveillance, robotics, human-machine interaction, motion analysis and so forth \cite{Li2020AsymmetricDC,Li2021LearningRC,LI2022108614}. Great progress has been witnessed in object tracking thanks to the successful application of deep learning to the field in recent years \cite{Jiao2021DeepLI}. Despite considerable progress in the field, current researches mainly focus on tracking generic objects, while very little attention is paid to tracking small and fast moving objects. Nevertheless, small and fast moving objects are common to see in the real world. Many of them are close connection to sports. Tracking of them is crucial to meet the practical and accuracy requirements of motion analysis for sports \cite{Colyer2018ARO,lapinski2019wide}, and to provide a fairer measure of performance than that provided by human judges, and to spare referees, umpires or judges from immense pressure in making accurate split-second decisions \cite{kerr2016technologies,Tamir2020TheMG}. It is also the key to develop automatic sports video recording and recommendation systems \cite{Jiang2021ResearchOM,Xue2017AutomaticVA}. However, tracking small and fast moving 
\begin{figure*}[h]
	\centering
	{
		\begin{minipage}[t]{1\textwidth}
			\centering
			\includegraphics[width=0.7\textwidth,height=0.15\textwidth]{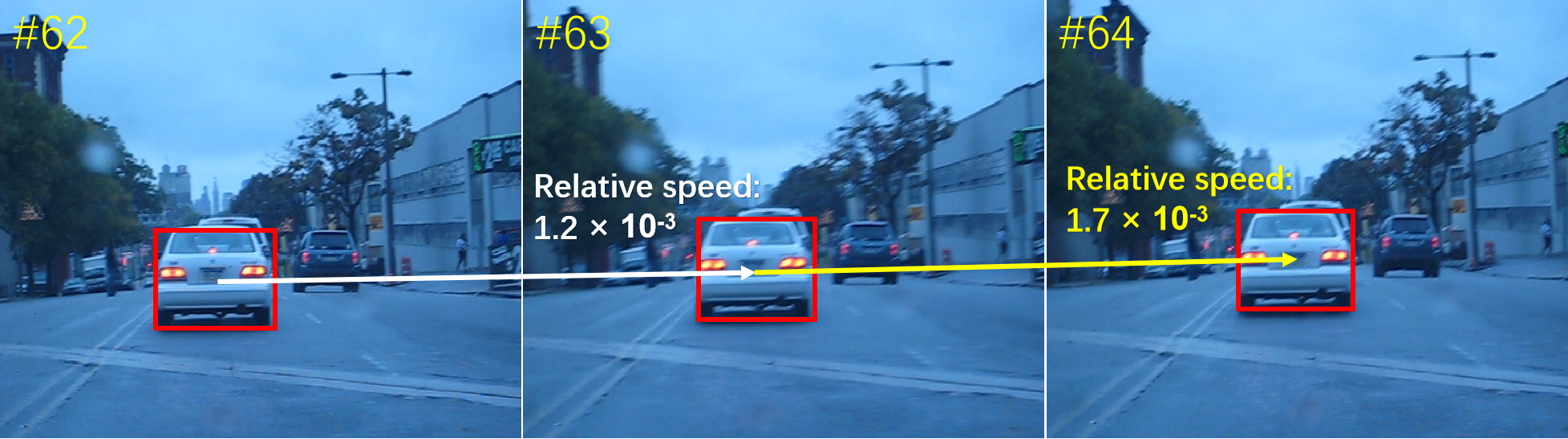}
			\centerline{\footnotesize (a) Example of generic object tracking.}
	\end{minipage}}
	{
		\begin{minipage}[t]{1\textwidth}
			\centering
			\includegraphics[width=0.7\textwidth,height=0.15\textwidth]{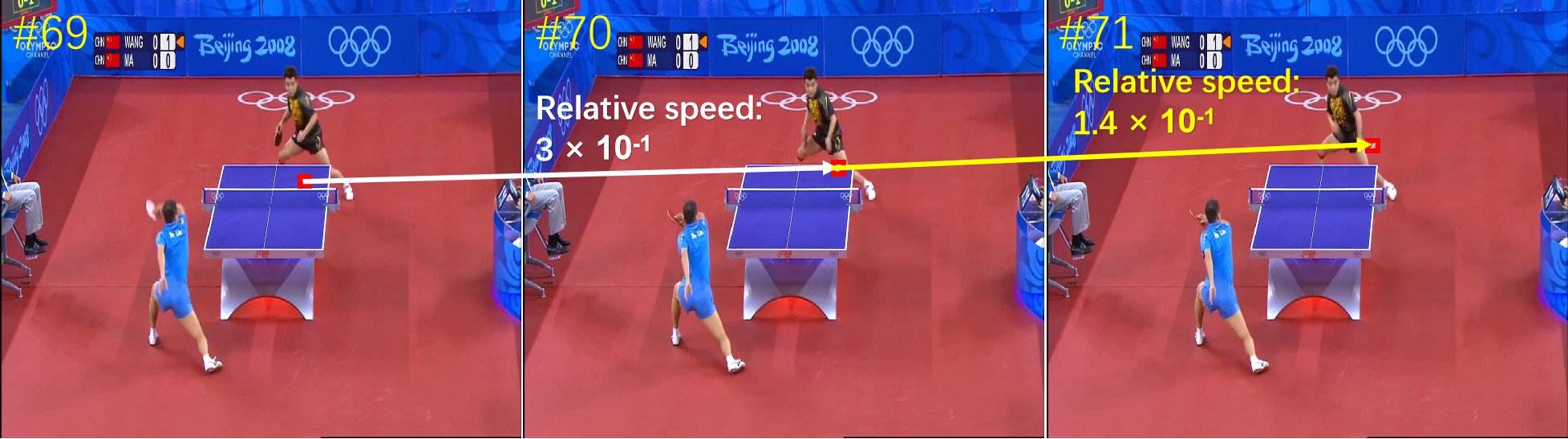}
			\centerline{\footnotesize (b) Example of small and fast moving object tracking.}
	\end{minipage}}
	\caption{Generic object tracking (a) and small and fast moving object tracking (b). Compared with generic object tracking, tracking of small and fast moving objects is more challenging as the objects are visually much smaller and the relative speeds are much higher. }
	\label{fig:TSFMO_examples}\vspace{-0.25in}
\end{figure*}
\begin{table*}
\centering
\caption{Comparison of average target size (in $pixel^2$) and average relative target speed (in 1/$pixel$) of object tracking benchmarks.}
\label{tab:dataset_target_size_speed}\vspace{-0.12in}
\resizebox{4.8in}{0.22in}{
\begin{tabular}{ccccccc} 
\toprule
                                      & OTB\cite{Wu2015ObjectTB}  & Got-10K\cite{Huang2021GOT10kAL} & LasoT\cite{Fan2019LaSOTAH} & TrackingNet\cite{Mller2018TrackingNetAL} & UAV123\cite{Mueller2016ABA} & \textbf{TSFMO (Ours)}  \\\hline
Avg. target size ($\times 10^3$)     & 6.73 & 228.97  & 56.17 & 45.65       & 2.48   & 0.51                   \\
Avg. relative speed ($\times 10^{-1}$) & 2.02 & 2.64    & 30.21 & 1.97        & 7.74   & 58.28                  \\
\bottomrule
\end{tabular}
}\vspace{-0.25in}
\end{table*}
objects are more challenging. There are several reasons account for this. First, small objects in sports are usually balls or objects of regular geometric shapes. Discriminative information provided by their shapes are fairly limited. Without sufficient discriminative cue of the target, existing tracking algorithms are prone to failure. Second, these objects are covered with plain or regular patterns, making the identifying of them in complex scenes very difficult, as they may be treated as parts of the background. 
Third, they are frequently moving at a high speed, which may cause severe motion blur in the captured images, and they may change moving direction abruptly at the moment of collision or being hit.


In addition to the above technical difficulties, another important reason that tracking small and fast moving objects is hardly touched is the lack of public available benchmarks, which are undoubtedly crucial to attack the problem and to advance the field as without which researchers are unable to effectively design and evaluate novel algorithms for improvement. Despite that there exist plenty of benchmarks for generic object tracking \cite{Wu2015ObjectTB,Liang2015EncodingCI,Kristan2016ANP,Galoogahi2017NeedFS,Valmadre2018LongtermTI,Mller2018TrackingNetAL,Huang2021GOT10kAL,Fan2019LaSOTAH}, there is no benchmark dedicated to tracking small and fast moving objects, especially small objects in sports. Although many of existing benchmarks consist of small and fast moving objects, the numbers of both sequences and object classes are very limited. Tracking algorithms developed to target these benchmarks are generic but not effective in dealing challenges posed by small and fast moving objects. To facilitate research on tracking small and fast moving objects, in this paper we present a dedicated dataset to serve as the testbed for fair evaluation and comparison.

\subsection{Contribution}
In this work, we make the first attempt to explore tracking small and fast moving objects by introducing the TSFMO benchmark for Tracking Small and Fast Moving Objects. TSFMO is made up of a diverse selection of 26 classes of sports with each containing multiple sequences. TSFMO consists of a total of 250 sequences and about 50k frames. Each sequence is manually annotated with axis-aligned bounding boxes with different attributes for performance evaluation and analysis. As far as we know, TSFMO is the first benchmark dedicated to the task of tracking small and fast moving objects, especially in sports. 
Fig. \ref{fig:TSFMO_examples} illustrates the differences between generic object tracking and small and fast moving object tracking. Compared with generic object tracking in which the target size is relatively larger and the relative speed (the displacement of the target in two neighbouring frames divided by the square root of the average area of the neighbouring bounding boxes) is relatively lower, tracking of small and fast moving objects is more challenging as the objects are visually very small and the relative speeds are much higher. A quantitative comparison of average target size and average relative target speed between four public generic object tracking benchmarks and 
five TSFMO is shown in Table \ref{tab:dataset_target_size_speed}.

In addition, in order to understand the performance of existing tracking algorithms and provide comparisons for future research on TSFMO, we extensively evaluated 20 state-of-the-art tracking algorithms on TSFMO. Meanwhile we conducted an in-depth analysis of the evaluation results and observed several surprising findings.
First, the tracking performances of existing trackers are much lower in TSFMO than in most public benchmarks for generic object tracking, which suggests that most existing tracking methods may overlook very important factors so that they are not effective in tracking small and fast moving objects. Second, not all latest trackers whose performances rank high on OTB \cite{Wu2015ObjectTB}, Got-10K \cite{Huang2021GOT10kAL}, LasoT \cite{Fan2019LaSOTAH}, and TrackingNet \cite{Mller2018TrackingNetAL} are highly-ranked on TSFMO, which suggests that the generalization ability of some existing trackers is questionable.  
So, as as an unexplored or less studied problem, improving the generalization ability of tracking algorithms is valuable and interesting. These above observations imply the need to develop tracking algorithms devoted to tracking small and fast moving objects, which may also stimulate more generalizable tracking algorithms in the future.

Last but not the least, we introduce a baseline tracker in order to facilitate the development of tracking algorithms on TSFMO. The proposed tracker is based on KeepTrack \cite{Mayer2021LearningTC} given that it shows the best performance among state-of-the-art trackers to be evaluated here. In view of that small objects may not have representation in deeper layers' features because of the larger receptive field in deeper layers and that combining low and high level features to boost performance has been extensively studied in tiny object detction \cite{Hong2022SSPNetSS,Kong2020FoveaBoxBA,Gong2021EffectiveFF}, we modify the architecture of KeepTrack \cite{Mayer2021LearningTC} so that low-level features are exploited to improve tracking performance. This results in the proposed tracker, which is called S-KeepTrack. 
The proposed S-KeepTrack outperforms all 20 state-of-the-art trackers on TSFMO.
In summary, we have made the following contributions:
\begin{itemize} 
    \item We propose TSFMO, which is, as far as we know, the first benchmark dedicated to track small and fast moving objects, especially in sports.
    \item We evaluate 20 state-of-the-art tracking algorithms with in-depth analysis to assess their performance and provide comparisons on TSFMO.
    \item We develop a baseline tracker S-KeepTrack base on the KeepTrack \cite{Mayer2021LearningTC} to encourage further research on TSFMO. 
\end{itemize}


\section{Related Works}
\subsection{Visual Tracking Algorithms}
Visual tracking has been studied for decades with a huge literature, the comprehensive review of which is out of scope of this paper. In this section, we review two popular trends including discriminative correlation filter (DCF)-based tracking and deep learning (DL)-based tracking in the field and refer readers
to \cite{Lu2019OnlineVT,Mazzeo2019VisualOT} for comprehensive surveys.

Roughly stated, DCF-based trackers treat visual tracking as an online regression problem. Thanks to the Parseval theorem and the Fast Fourier Transform (FFT), DCF-based tracker can be effectively evaluated in the frequency domain and demonstrate impressive CPU speeds \cite{li2021learning}.
They started with the minimum output sum of squared error (MOSSE) filter \cite{Bolme2010VisualOT}. After that great advance has been witnessed in DCF-based trackers \cite{li2021learning}. 
For instance, an additional scale filter is exploited in \cite{li2014a,danelljan2017discriminative} to deal with target scale variations. The trackers in \cite{danelljan2015learning,li2018learning} leverage regularization techniques to improve robustness. The approach in \cite{Li2020AsymmetricDC} generalize the DCF to achieve translation equivariance. The methods in \cite{Ma2015HierarchicalCF,Danelljan2016BeyondCF} utilize deep features instead of handcrafted ones in correlation filter tracking and achieve significant improvements.

The great success of deep learning in other vision tasks motivated the mushrooming development of DL-based trackers in visual tracking in recent years. As one of the pioneering works, SiamFC \cite{bertinetto2016fully} considered visual tracking as a general similarity-learning problem and took advantage of the Siamese network \cite{chicco2021siamese} to measure the similarity between target and search image. Since then, many DL-based trackers base on Siamese architectures have been proposed \cite{li2019siamrpn++,guo2020siamcar,xu2020siamfc++} and the tracking performances have been significantly improved.
Along another line, visual tracking is divided into two sub-tasks, i.e., localization and scale estimation, which are solved, respectively, by an online classifier and an offline intersection-over-union (IoU) network \cite{Danelljan2019ATOMAT,bhat2019learning,danelljan2020probabilistic,Mayer2021LearningTC}. 

\subsection{Visual Tracking Benchmarks}
As standards by which the performances of tracking methods are measured or judged, tracking benchmarks, undoubtedly, are crucial for the development of visual tracking. 
Existing benchmarks can be roughly divided into two types: generic benchmarks and specific benchmarks \cite{HengFan2020TransparentOT}.

\textbf{Generic Benchmarks.} A generic tracking benchmark is usually designed for tracking objects in general scenes. OTB-2013 \cite{Wu2015ObjectTB}
is the first generic benchmark with 50 sequences and later extended to OTB-2015 with 100 sequences. TC-128 \cite{Liang2015EncodingCI} consists of 128 colorful sequences, and is used to study the impact of color information on tracking performance. VOT \cite{Kristan2016ANP} organizes a series of tracking competitions with up to 60 sequences. NfS \cite{Galoogahi2017NeedFS} concerns about videos of high frame rate. NUS-PRO \cite{Li2016NUSPROAN} collects 365 videos and primarily addresses tracking rigid objects. TracKlinic \cite{Fan2021TracKlinicDO} includes 2,390 videos to evaluate tracking algorithms under various challenges. Recently, many large-scale benchmarks have been proposed to provide training data for developing DL-based trackers. OxUvA \cite{Valmadre2018LongtermTI} provides 366 videos aiming for long-term traking in the wild. TrackingNet \cite{Mller2018TrackingNetAL} collects a large-scale dataset consisting of more than 30K sequences for deep tracking. GOT-10k \cite{Huang2021GOT10kAL} provides 10K sequences with rich motion trajectories.
LaSOT offers 1,400 long-term videos in \cite{Fan2019LaSOTAH} and later introduces additional 150 sequences and a new evaluation protocol for unseen objects in \cite{Fan2021LaSOTAH}.

\textbf{Specific Benchmarks.} In addition to generic visual tracking benchmarks, there exist specific benchmarks for particular goals. UAV123 \cite{Mueller2016ABA} consists of 123 sequences captured by unmanned aerial vehicle (UAV) for low altitude UAV target tracking. VOT-TIR \cite{Kristan2017TheVO} is from VOT and focuses on object tacking in RGB-T sequences, aiming at taking advantage of RGB and thermal infrared images simultaneously. CDTB \cite{Lukei2019CDTBAC} and PTB \cite{Song2013TrackingRU} are designed to assess tracking performance on RGB-D videos, D indicating depth images. TOTB \cite{HengFan2020TransparentOT} collects 225 videos from 15 transparent object categories and focus on transparent object tracking.

Despite of the availability of the above benchmarks, they mainly focus on tracking objects of common sizes and relatively slow speeds. Tracking of small and fast moving objects, especially in sports, has received very little attention. The most important reason, we think, is the lack of public available benchmarks, which motivates our proposal of TSFMO.

\subsection{Dealing with Small and Fast Moving Objects in Vision}
Small objects here refer to objects with smaller physical sizes in the real world and occupying areas less than and equal to 32 × 32 pixels \cite{Tong2020RecentAI}, while fast moving objects refer to the ones that may move over a distance exceeding its size within the exposure time \cite{Rozumnyi2017TheWO}. Small and fast moving objects are common to see in the real-world, and a significant amount of researches have been devoted to deal with them.
For example, the methods of \cite{Kembhavi2011VehicleDU,Morariu2014CompositeDF} studied the problem of small object detection utilizing hand-engineered features and shallow classifiers in aerial images.
The approach in \cite{Huang2008AMO} combined detection and tracking and integrated them into an adaptive prticle filter to handle small object localization, but it was evaluated on mere two testing videos for case study.
To the best of our knowledge, Chen et al. \cite{Chen2016RCNNFS} are perhaps the first to introduce a small object detection (SOD) dataset, an evaluation metric, and provide
a baseline score in order to explore small object detection.
The work of \cite{Ahmadi2016SmallDO} presented an algorithm for detecting
and tracking small dim targets in Infrared (IR) image sequences base on the frequency and spatial domain information.
The work of \cite{Rozumnyi2017TheWO} presented a method for detecting and tracking fast moving objects and provided a new dataset consisting of 16 sports videos for evaluation. In \cite{Liu2020AggregationSF}, an aggregation signature was proposed for small object tracking and 112 sequences were collected for evaluation. The method in \cite{Zita2021TrackingFM} implemented a segmentation network that performs near real-time detection and tracking of fast moving objects and introduced a synthetic physically plausible fast moving object sequence generator for training purpose. 
The method of \cite{Zaveri2004SmallAF} investigated the problem of small and fast moving object detection and tracking in sports video sequences, using only motion as a cue for detection and multiple filter banks for tracking.

Our work is related to \cite{Huang2008AMO,Ahmadi2016SmallDO,Rozumnyi2017TheWO,Liu2020AggregationSF,Zita2021TrackingFM,Zaveri2004SmallAF} but different in: (1)
TSFMO focuses on tracking small and fast moving objects, while other works concentrate on either
tracking small objects \cite{Huang2008AMO,Ahmadi2016SmallDO,Liu2020AggregationSF}, or tracking fast moving objects \cite{Rozumnyi2017TheWO,Zita2021TrackingFM}. (2) Although both focus on tracking small and fast moving objects, TSFMO provides a diverse benchmark of hundreds challenging sequences for evaluation, while in \cite{Zaveri2004SmallAF} only a small number of sequences is provided, which are captured with a stationary camera under indoor conditions where the background
remains stationary. Such data is way far from real applications. 

\section{Tracking Small and Fast Moving Objects}


\subsection{Video Collection}

We select 26 small and fast-moving object categories to construct TSFMO, including  baminton, baseball, basketball, beach volleyball, bowling, boxing, curling, discus, football, gateball, golf, hammer, handball, ice hockey, indoor football, kick volleyball, pingpong, polo, ruby, shot, shuttlecock ball, snooker, squash, tennis, volleyball and water polo
. Fig. \ref{fig:datasetSamples} demonstrates some sample sequences from these categories.

\begin{figure*}[h]
	\centering
	\includegraphics[width=0.6\textwidth]{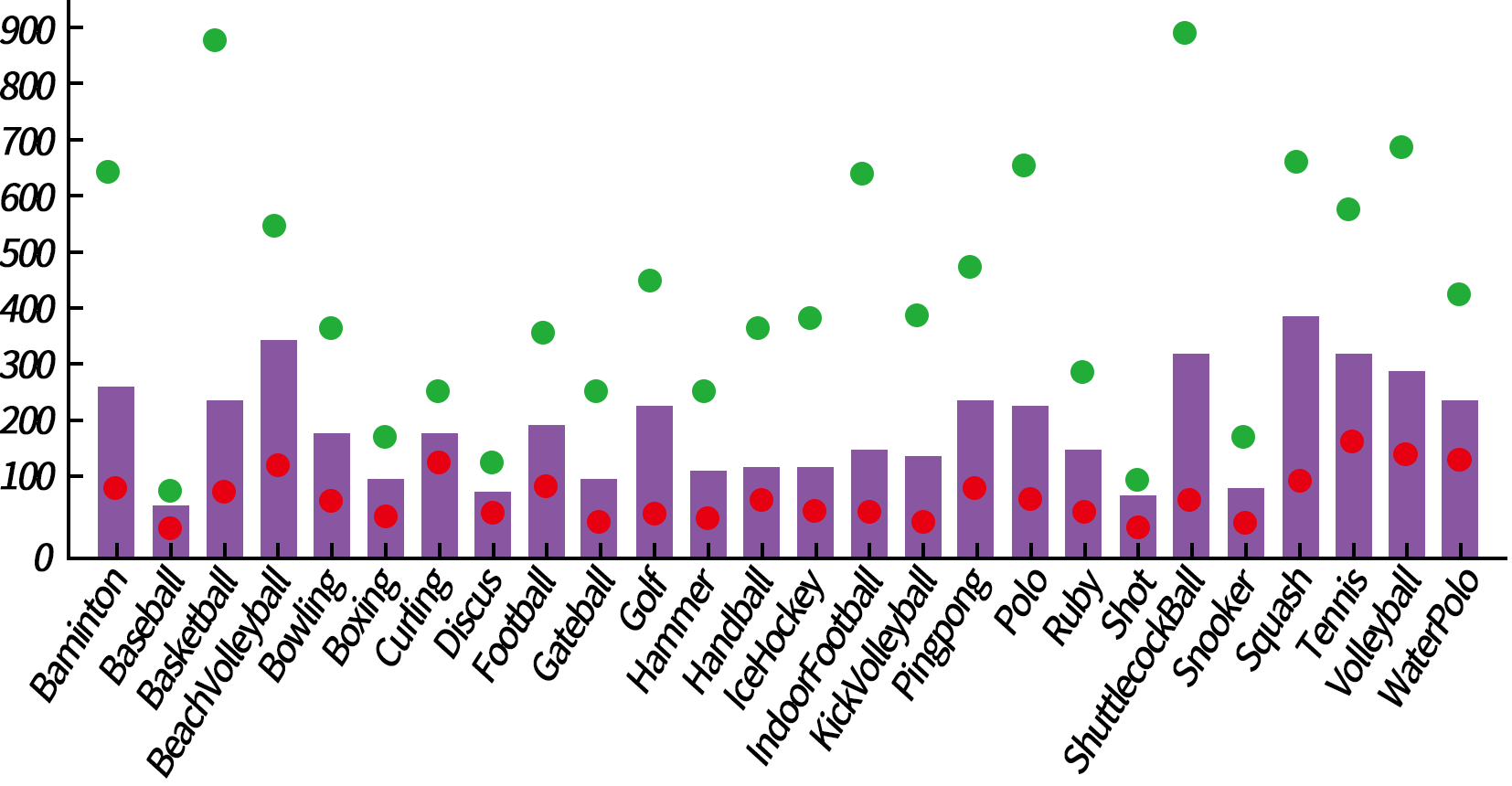}	\vspace{-0.15in}
	\caption{Average video length for each object class in TSFMO. The red and green dots indicate the minimum and maximum frame numbers of each category.}
	\label{fig:avg_length}	
\end{figure*}

After determining the object categories, we search for raw sequences of each class from the Internet, as it is the source of many tracking benchmarks (e.g., LaSOT \cite{Fan2021LaSOTAH}, GOT-10k \cite{Huang2021GOT10kAL}, TrackingNet \cite{Mller2018TrackingNetAL}, etc). 
Initially, we collected at least 10 raw videos for each class and gathered more than 280 sequences in total. We then carefully inspect each sequence for its availability for tracking and drop the undesirable sequences.
Afterwards, we verify the content of each raw sequence and remove the irrelevant parts to obtain a video clip that is suitable for tracking. 
We intentionally limit the number of frames in each video to 900 frames, which is enough for accessing the tracker's performance on tracking small and fast-moving objects, meanwhile manageable for annotation.
Eventually, TSFMO is made up of 250 sequences from 26 object classes with about 50K frames. Table \ref{tab:TSFMO_statistic} summarizes TSFMO, and Fig. \ref{fig:avg_length} shows the average sequence length for each object category in TSFMO.

\subsection{Annotation}

For sequence annotation, we follow the principle used in [14]: given an initial object, for each frame in the sequence, the annotator draws/edits an axis-aligned 
\vspace{-0.45in}
\begin{table*}[h]
	\centering
	\caption{ Summary of statistics of the proposed TSFMO.}
	\label{tab:TSFMO_statistic}
	\vspace{-0.1in}
	\resizebox{4in}{0.2in}{
		\begin{tabular}{ll|ll|ll|ll} 
			\toprule
			Number of videos & 250 & Min frames & 16  & Frame rate        & $\leq$30fps & Max frames   & 887   \\
			Total frames     & 49k & Avg frames & 196 & Object categories & 26          & Avg duration & 7.4s  \\
			\bottomrule
		\end{tabular}
	}\vspace{-0.15in}
\end{table*}
bounding box as the tightest bound box to fit any visible part of the object if the object appears; otherwise, either an absence label or out of view (OV) or full occlusion (FOC) is assigned to the frame.

\begin{figure*}[t]
	\centering
	\includegraphics[width=1\textwidth]{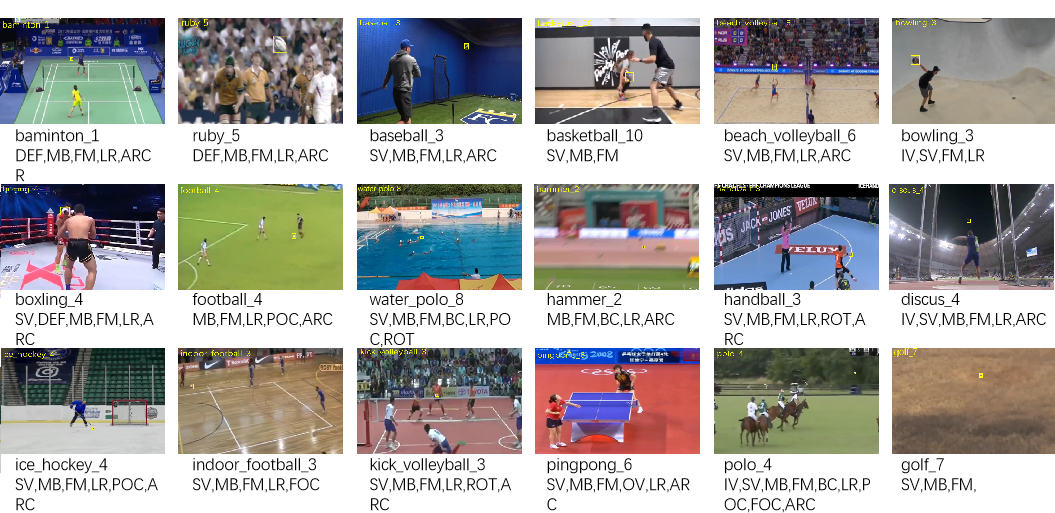}	\vspace{-0.25in}
	\caption{Example sequences of small and fast moving object tracking in our TSFMO. Each sequence is annotated with axis-aligned bounding boxes and attribues.}
	\label{fig:datasetSamples}
\end{figure*}

Adhering to the above principle, we finish the annotation in three steps, i.e., manual annotation, visual inspection, and box refinement. In the first step, each video was labelled by an expert, i.e., a student working on tracking. As annotation errors or inconsistencies is hardly avoidable in the first stage, a visual inspection is performed to verify the annotations in the second stage, which is conducted by a validation team. If the validation team do not agree on the annotation unanimously, it will be sent back to the original annotator for refinement in the third step. This three-step strategy ensures high-quality annotation for objects in TSFMO. See Fig. \ref{fig:datasetSamples} for some examples of box annotations for TSFMO.

\subsection{Attributes}
In view of that in-depth analysis of tracking methods is crucial to grasp their strengths and limitations, we select twelve attributes that widely exist in video tasks and annotate each sequence with these attributes, including (1) Illumination Variation (IV), (2) Deformation (DEF), (3) Motion Blur (MB), (4) Rotation (ROT), (5) Background Clutter (BC), (6) Scale Variation (SV),  which is assigned when the ratio of bounding box is outside the range [0.5, 2],  (7) Out-of-view (OV), (8) Low Resolution (LR), which is assigned when the target area is smaller than 900 pixels, (9) Aspect Ratio Change (ARC), which is assigned when the ratio of the bounding box aspect ratio is outside the range [0.5, 2], (10) Partial Occlusion (POC), (11) Full Occlusion (FOC), and (12) Fast Motion (FM), which is assigned when the target center moves by at least 50\% of its size in last frame. 
Table \ref{tab_attribute_dist} shows the distribution of these attributes on TSFMO. 
\begin{table*}[h]
	\centering
	\caption{Distribution of twelve attributes on the TSFMO. The diagonal (shown in \textbf{bold}) corresponds to the distribution over the entire benchmark, and each row or column presents the joint distribution for the attribute subset.}
	\label{tab_attribute_dist}\vspace{-0.1in}
	\resizebox{3in}{0.62in}{
		\begin{tabular}{@{}ccccccccccccc@{}}
			\toprule
			& IV                           & SV                            & DEF                          & MB                            & FM                            & OV                           & BC                           & LR                            & POC                          & ROT                           & FOC                         & ARC                           \\ \hline\hline
			IV  & \textbf{31} & 29                            & 0                            & 26                            & 31                            & 0                            & 14                           & 30                            & 7                            & 12                            & 1                           & 23                            \\
			SV  & 29                           & \textbf{193} & 33                           & 170                           & 191                           & 17                           & 49                           & 188                           & 53                           & 96                            & 8                           & 147                           \\
			DEF & 0                            & 33                            & \textbf{45} & 39                            & 44                            & 12                           & 5                            & 38                            & 7                            & 26                            & 0                           & 41                            \\
			MB  & 26                           & 170                           & 39                           & \textbf{221} & 132                           & 18                           & 58                           & 203                           & 56                           & 102                           & 9                           & 158                           \\
			FM  & 31                           & 191                           & 44                           & 132                           & \textbf{248} & 18                           & 58                           & 231                           & 56                           & 102                           & 9                           & 75                            \\
			OV  & 0                            & 17                            & 12                           & 18                            & 18                            & \textbf{21} & 0                            & 20                            & 4                            & 17                            & 0                           & 17                            \\
			BC  & 14                           & 49                            & 5                            & 58                            & 58                            & 0                            & \textbf{63} & 59                            & 18                           & 21                            & 1                           & 40                            \\
			LR  & 30                           & 188                           & 38                           & 203                           & 231                           & 20                           & 59                           & \textbf{235} & 62                           & 108                           & 8                           & 170                           \\
			POC & 7                            & 53                            & 7                            & 56                            & 56                            & 4                            & 18                           & 62                            & \textbf{71} & 31                            & 1                           & 46                            \\
			ROT & 12                           & 96                            & 26                           & 102                           & 102                           & 17                           & 21                           & 108                           & 31                           & \textbf{111} & 2                           & 80                            \\
			FOC & 1                            & 8                             & 0                            & 9                             & 9                             & 0                            & 1                            & 8                             & 1                            & 2                             & \textbf{9} & 7                             \\
			ARC & 23                           & 147                           & 41                           & 158                           & 75                            & 17                           & 40                           & 170                           & 46                           & 80                            & 7                           & \textbf{176} \\ \bottomrule
		\end{tabular}
	}	\vspace{-0.12in}
\end{table*}
As can be seen, the most common challenge in TSFMO is Fast Motion. In addition, the Motion Blur and Low Resolution also present frequently in TSFMO.

\section{A new baseline : S-KeepTrack}
We found that among the state-of-the-art trackers to be evaluated here KeepTrack shows the best performance, despite that it is still far from satisfactory. To facilitate the development of tracking algorithms for tracking small and fast moving objects, we present a new baseline tracker based on the KeepTrack. The proposed tracker, dubbed S-KeepTrack, combines low-level and high-level features to improve tracking performance, considering that small objects may not have representation in deeper layers' features because of their larger receptive field. In contrast to one target candidate association network of KeepTrack, S-KeepTrack has two parallel target candidate association networks that separately process the feature encodings of lower and higher level features, respectively. And 
\begin{figure*}[t]
	\centering
	\includegraphics[width=1\textwidth,height=0.4\textwidth]{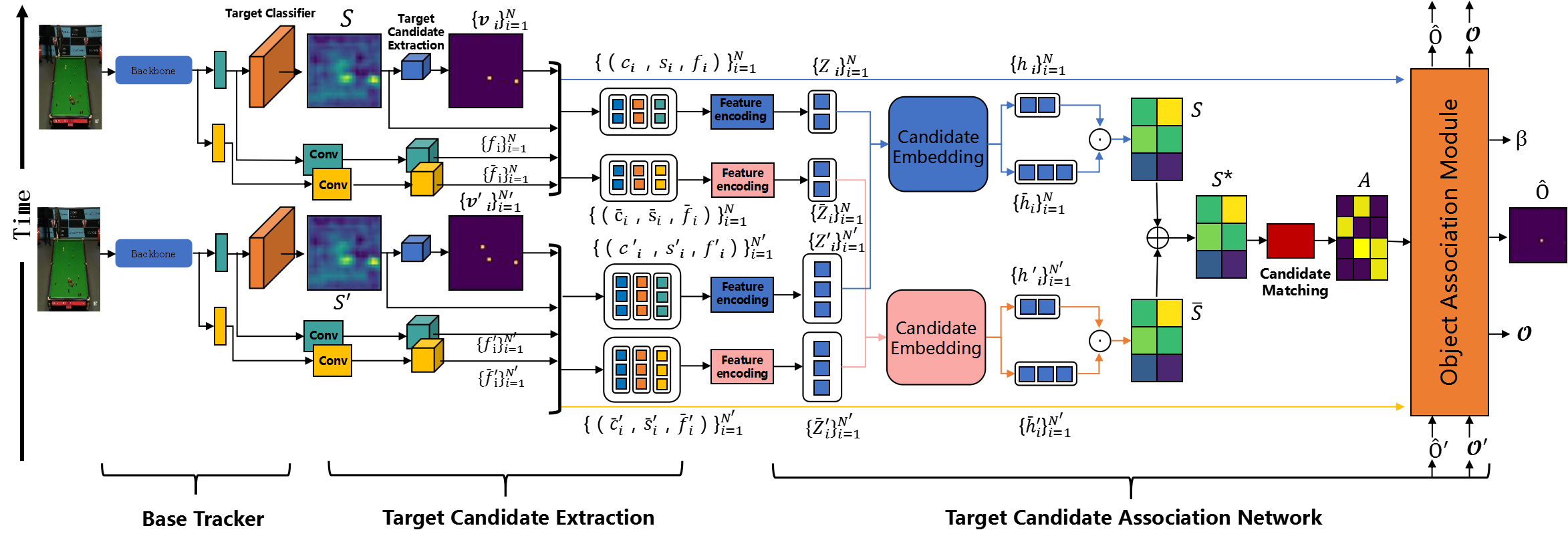}\vspace{-0.15in}
	\caption{Overview of the tracking pipline of the proposed S-KeepTrack. Note that the parallel architectures plotted in the same color share parameters.} 
	\label{S-KeepTrack_overview}
\end{figure*}
the result of candidate association is a weighted combination of the result of two parallel networks. Hopefully, this combination of low-level and high-level features will build a stronger tracker than KeepTrack for tracking small and fast moving objects.

Like KeepTrack, S-KeepTrack also consists of three components: i) a base tracker that predicts the target score map ${s}$ for the current frame and extracts the target candidates $V$ and $V'$ ($V=\{v_i\}_{i=1}^{N}$ and $V'=\{v'_i\}_{i=1}^{N'}$ denote candidate set of the current and the previous frame, respectively) by finding locations in ${s}$ with high target score, ii) a target candidate extraction module that extracts for each candidatea a set of features (i.e., target classifier score $s_i$, location $c_i$ in the image, and two appearance cues $\bar{f}_i$ and $f_i$ from lower and higher level, respectively, where $i$ indexes the $i$th candidate), and iii) a target candidate association network that estimates the candidate assignment probabilities between two consecutive frames. The difference between S-KeepTrack and KeepTrack lies in the parallel network achitechtures designed for the low level features from the backbone. An overview of our tracking pipeline is shown in Fig. \ref{S-KeepTrack_overview}.

\noindent\textbf{Base tracker:} The base tracker is inherited from KeepTrack with the difference: the backbone of our S-KeepTrack outputs both low and high level features for target candidate extraction instead of only high level one as in KeepTrack.

\noindent\textbf{Target candidate extraction:} This module aims to build a feature representation for each target candidate. In KeepTrack, the position $c_i$ and the target classifier score $s_i$ are taken as two discriminative cues of the target candidate $v_i$, based on two observations: i) the motion of the same object from frame to frame is typically small and thus the same object has similar locations in two neighbouring frames, ii) only small
changes in appearance for each object \cite{Mayer2021LearningTC}. In addition, it also processes the backbone features with a single learnable convolution layer to add a more discriminative appearance-based feature $f_i$. Finally, each feature tuple $(c_i, s_i, f_i)$ representing the target candidate $v_i$ is fed into a feature encoding module to get the code \cite{Mayer2021LearningTC}
\begin{equation}
\small
	z_i=f_i+\psi (s_i,c_i),
\end{equation} 
where $\psi$ denotes a Multi-Layer Perceptron (MLP), mapping $s_i$ and $c_i$ to the same dimensional space as $f_i$. $z_i$ then passes through a candidate embeding network before candidate matching is conducted.
In order to build a better feature representaion for each target candidate, we additionally process a low-level backbone feature with an extra learnable convolution layer in our S-KeepTrack, resulting in an extra discriminative appearance-based feature $\bar{f}_i$. In view of that the feature encoding module and the candidate embedding network have been carefully tailored to the tuple $(c_i, s_i, f_i)$, we avoid fusing $f_i$ and $\bar{f}_i$ to get an entangled feature, as it may not fit well with the original network and will demand modifying the feature encoding module and the candidate embedding network. Instead, from the perspective of ensemble method \cite{Zhou2012EnsembleMF}, we build a new tuple $(c_i, s_i, \bar{f}_i)$ to accompany $(c_i, s_i, f_i)$. And $(c_i, s_i, \bar{f}_i)$ is fed into a new feature encoding module to get another code
\begin{equation}
\small
	\bar{z}_i=\bar{f}_i+{\psi}' (s_i,c_i),
\end{equation} 
where ${\psi}'$ denotes a MLP that map $s_i$ and $c_i$ to the same dimensional space as $\bar{f}_i$. The feature encoding modules are followed by two parallel candidate embedding networks, which will be detailed in the following.

\noindent\textbf{Candidate embedding network:} 
On an abstract level, candidate association bares similarities with the task of sparse feature matching \cite{Sarlin2020SuperGlueLF,Mayer2021LearningTC}, for which KeepTrack adopted the SuperGlue \cite{Sarlin2020SuperGlueLF} architecture that establishes state-of-the-art sparse feature matching performance to do candidate embedding and matching. With this method, the feature encodings $\{z_i\}_{i=1}^{N}$ and $\{z'_i\}_{i=1}^{N}$ of two neighbouring frames translate to nodes of a single complete graph with two types of directed edges: 1) self edges within the same frame and 2) cross edges connecting only nodes between the frames. In SuperGlue \cite{Sarlin2020SuperGlueLF}, a Graph Neural Network (GNN) is utilized to send messages in an alternating fashion across self or cross edges to produce a new feature representation for each node after every layer, in which self and cross attention are used to compute the messages for self and cross edges \cite{Sarlin2020SuperGlueLF,Mayer2021LearningTC}. After the last message passing layer a linear projection layer extracts the final feature representation $h_i$ for each candidate $v_i$ \cite{Mayer2021LearningTC}. In our S-KeepTrack, a new candidate embedding network is adopted to deal with the feature encoding related to the low-level features, resulting in an extra feature representation $\bar{h}_i$ for each candidate $v_i$.

\noindent\textbf{Candiate matching:} 
In KeepTrack, the candidate embeddings $h'_i$ and $h_j$ (corresponding to two candidates $v'_i \in V'$ and
$v_j \in V$, respectively.) are used to compute the similarity between $v'_i$ and $v_j$ by the scalar product: $S_{i,j}=\left \langle  h'_i,h_j \right \rangle$.
Given a match may not exist for every candidate, KeepTrack makes use of the dustbin concept \cite{DeTone2018SuperPointSI,Sarlin2020SuperGlueLF} to actively match candidates that miss their counterparts to the so-called dustbin, ending up with an augmented assignment matrix $A$ with an additional row and column representing dustbins. Note that a dustbin is a virtual candidate without any feature representation, to which a candiate corresponds only if its similarity scores to all other candidates are sufficiently low. To obtain the assignment matrix $A$ between $V$ and $V'$ given the similarity matrix $S=\{S_{i,j}\}$, KeepTrack follow Sarlin et al. \cite{Sarlin2020SuperGlueLF} and designed a learnable module to predict $A$. However, we have two parallel similarity matrice $S$ and $\bar{S}$ in S-KeepTrack because of the parallel feature representations. Therefore, we aggregate the two similarity matrice $S$ and $\bar{S}$ to produce a fused one $S^*$ by the following weighted sum,
\begin{equation}\label{Eq:weight_coefficient}
\small
	S^*=\omega S + (1-\omega)\bar{S},
\end{equation} 
where $\omega \in [0,1]$ is the weight coefficient to balance the contributions of $S$ and $\bar{S}$. $S^*$ is then fed into the candidate matching module to predict the assignment matrix $A$ as in KeepTrack. 

\noindent\textbf{Object association:} 
The object association module uses the estimated assignments to determine the object correspondences during online tracking, which follows that of KeepTrack. The idea is to keep track of every object present in each scene over time using a database $\mathcal{O}$ with each entry being an object visible in the current frame. When online tracking, the estimated assignment matrix $A$ is used to determine which objects disappeared, newly appeard, or stayed visible and can be associated unambiguoursly, and to help reason the target object ${\hat{o}}$. Last but not least, the target detection confidence $\beta$ is computed to manage the memory and control the sample wieght for updating the target classifier online. This finishes the description of our S-KeepTrack. It is worth noting that the losses and training pipeline of S-KeepTrack is the same as KeepTrack. Please refer to \cite{Mayer2021LearningTC} for details.

\section{Evaluation}

\subsection{Evaluation Metrics} 

We use one-pass evaluation (OPE) and measure each tracker using precision, and success rate as in \cite{Fan2019LaSOTAH,Mller2018TrackingNetAL}. The precision measures the distance between the centers of the estimated target bounding box and the groundtruth box in pixels. 
Success rate is based on the intersection over union (IoU)
of the estimated target bounding box and the groundtruth box, specifically, it measures the percentage of estimated target bounding boxes with IoU larger than 0.5 \cite{Wu2015ObjectTB,Fan2019LaSOTAH,Mller2018TrackingNetAL}. The precision at 20 pixels (PRC) and the area under curve (AUC) of success plot is usually used for ranking.

\subsection{Trackers for Comparison} 

%


We evaluate 20 state-of-the-art trackers to understand their performance on TSFMO, including KeepTrack \cite{Mayer2021LearningTC}, AutoMatch \cite{Zhang2021LearnTM}, TransT \cite{Chen2021TransformerT}, SAOT \cite{Zhou2021SaliencyAssociatedOT}, SiamGAT \cite{Guo2021GraphAT}, LightTrack \cite{Yan2021LightTrackFL}, TrDiMP \cite{Wang2021TransformerMT}, TrSiam \cite{Wang2021TransformerMT}, KYS \cite{Bhat2020KnowYS},
HIFT \cite{Cao2021HiFTHF}, SuperDiMP \cite{Kristan2020TheEV}, PrDiMP50 \cite{danelljan2020probabilistic}, PrDiMP18 \cite{danelljan2020probabilistic}, Ocean \cite{Zhang2020OceanOA}, SiamMask \cite{Wang2019FastOO},
 SiamRPN++ \cite{li2019siamrpn++}, ATOM \cite{Danelljan2019ATOMAT}, DiMP50 \cite{bhat2019learning}, DiMP18 \cite{bhat2019learning}, and SiamDW \cite{Zhang2019DeeperAW}. 

\begin{figure*}[h]
	\centering
	\subfigure{
		\begin{minipage}[t]{0.48\textwidth}
			\includegraphics[width=1\textwidth, height=0.58\textwidth]{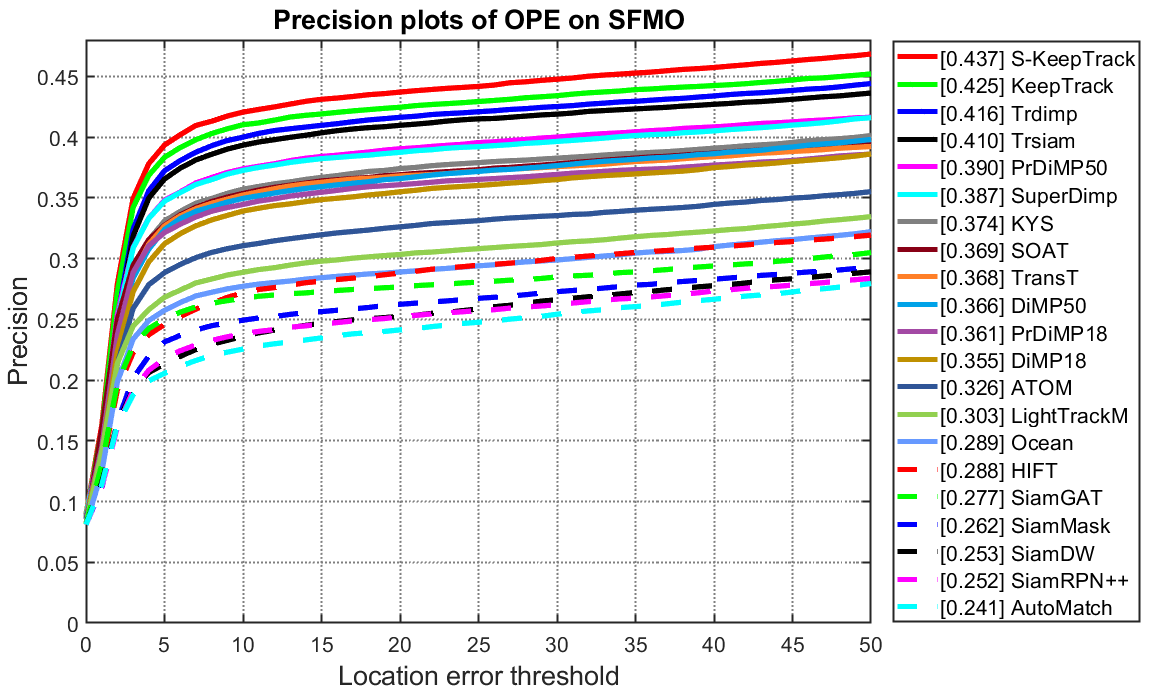}\hspace{0in}
	\end{minipage}}
	\subfigure{
		\begin{minipage}[t]{0.48\textwidth}
			\includegraphics[width=1\textwidth, height=0.58\textwidth]{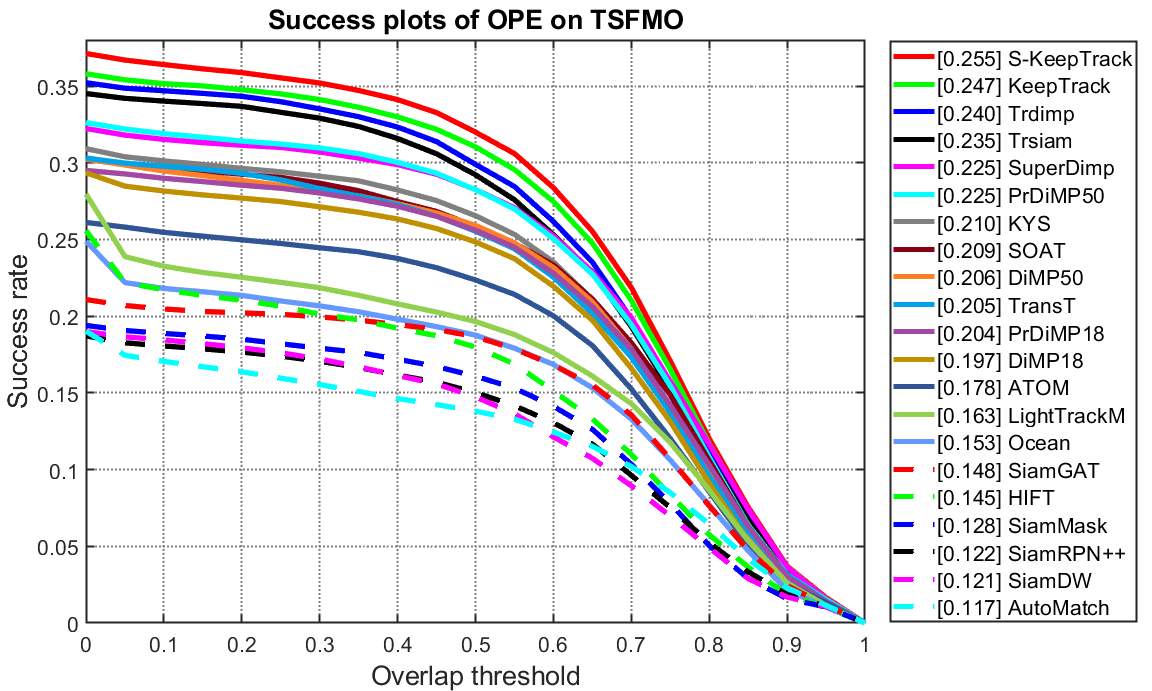}\hspace{0in}
	\end{minipage}}	\vspace{-0.15in}
	\caption{Overall performance on TSFMO. Precision and success rate for one-pass evaluation (OPE) \cite{2013Online} are used for evaluation. 
	}	\vspace{-0.15in}
	\label{fig_overall_p_s_plots}
\end{figure*}
\begin{figure*}[h]
	\centering
	\subfigure{
		\begin{minipage}[t]{0.48\textwidth}
			\includegraphics[width=1\textwidth,height=0.58\textwidth]{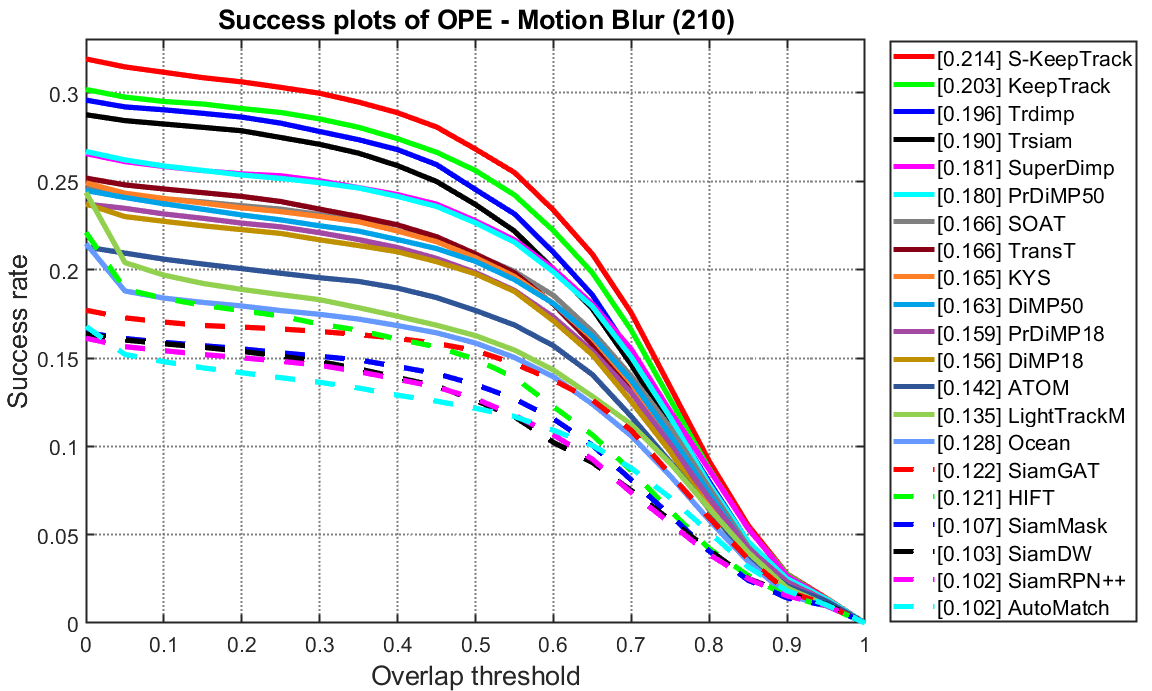}\hspace{0in}
	\end{minipage}}
	\subfigure{
		\begin{minipage}[t]{0.48\textwidth}
			\includegraphics[width=1\textwidth, height=0.58\textwidth]{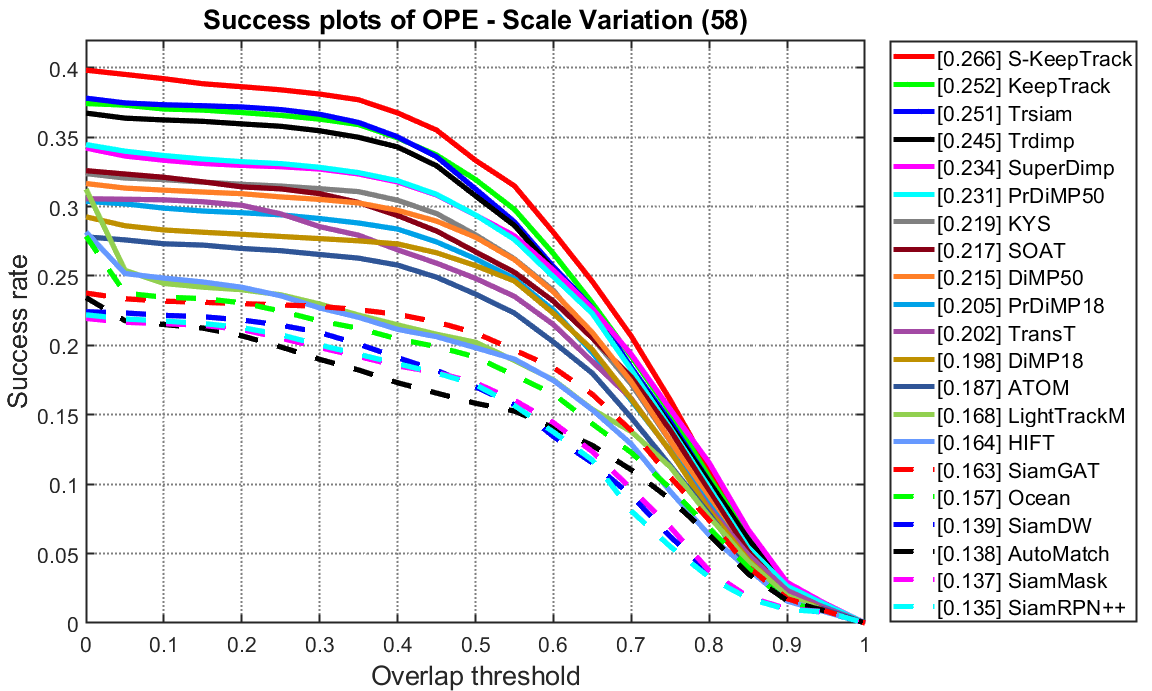}\hspace{0in}
	\end{minipage}}	\vspace{-0.15in}
	\caption{Attribute-based comparison on motion blur and scale variation.}
	\label{fig_attribute_p_s_plots}	\vspace{-0.15in}
\end{figure*}
\subsection{Evaluation results}

\textbf{Overall performance.} 20 state-of-the-art trackers and our S-KeepTrack are extensively evaluated on TSFMO. Note that existing trackers are used without any modification. 
The evaluation results are reported in precision and success plot, as shown in Fig. \ref{fig_overall_p_s_plots}. As can be seen, our S-KeepTrack achieved the best results with a PRC of 0.437, AUC of 0.255.  KeepTrack got the second best PRC of 0.425, and likewise KeepTrack got the second best AUC of 0.247. In comparison with the second best tracker KeepTrack, S-KeepTrack achiveves improvements of 1.2\% and 0.8\% in terms of PRC and AUC, respectively, which evidences the effectiveness and advantage
of our method of combining low level and high level features for small and fast moving object tracking.

\textbf{Attribute-based performance.} We conduct performance evaluation under twelve attributes to further analyze
and understand the performances of different trackers. Our S-KeepTrack achieves the best PRC and AUC on most attributes. Due to space limitation, we demonstrate in Fig. \ref{fig_attribute_p_s_plots} the success plots for the two most frequent challenges, including motion blur and scale variation. 
We observe that S-KeepTrack performs the best on both attributes. Specifically, S-KeepTrack achieves a AUC of 0.214 on motion blur, surpassing the second best tracker KeepTrack with AUC of 0.203 by 1.1\%; on scale variation, S-KeepTrack' AUC is 0.266, outperforming the second best tracker KeepTrack with AUC of 0.252 by 1.4\%. This also supports the importance of combining low level and high level features for small and fast moving object tracking.

\begin{figure*}[t]
	\centering
	\includegraphics[width=1\textwidth,height=0.51\textwidth]{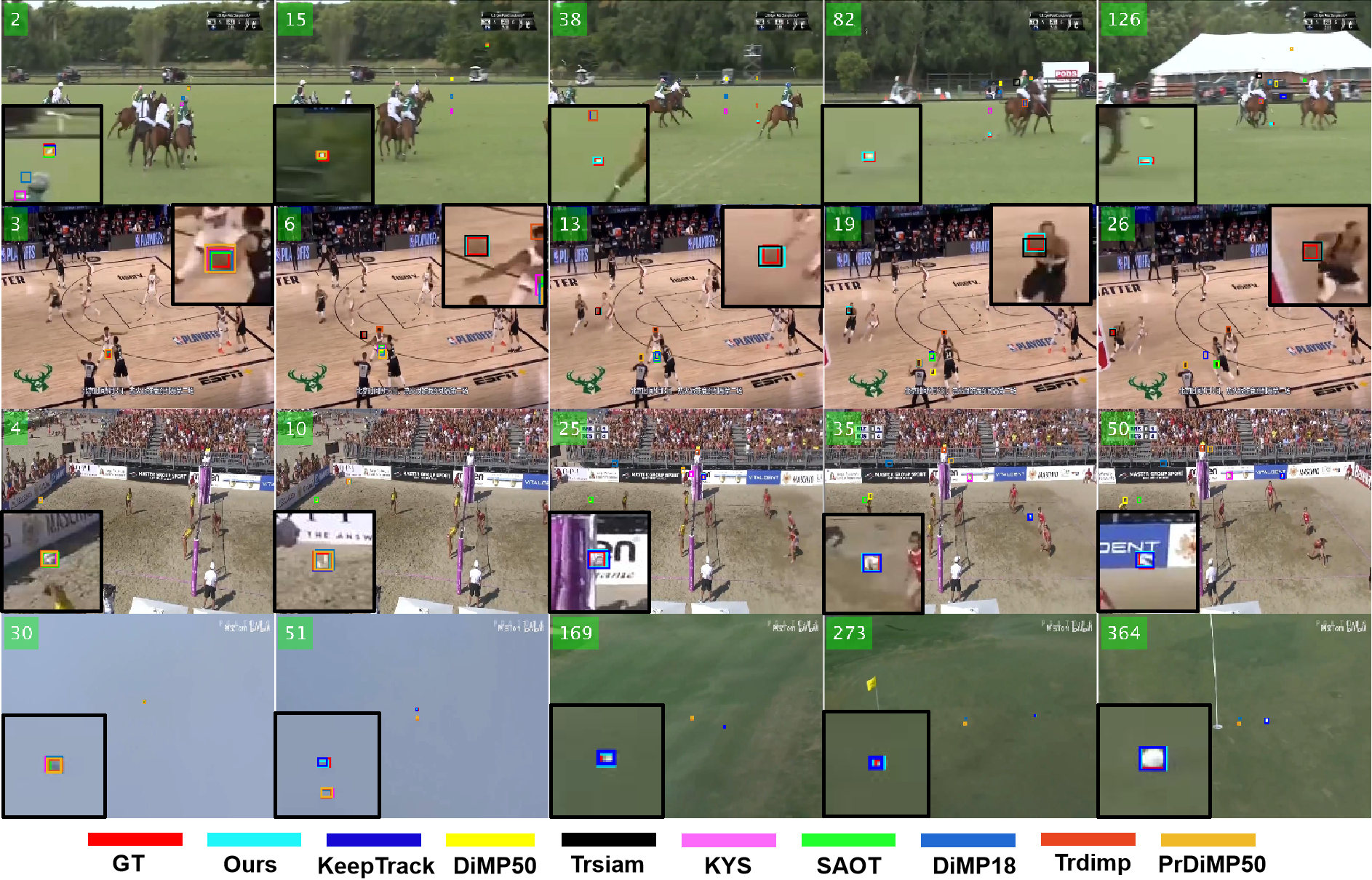}\vspace{-0.1in}
	\caption{Qualitative evaluation on 4 sequences from TSFMO, i.e., polo\_4, basketball\_14, beach\_volleyball\_8, and golf\_4 from top to bottom. The results of different methods have been shown with different colors, and 'GT' denotes the groundtruth. } \label{fig:visual_examples}\vspace{-0.15in}
\end{figure*}

\textbf{Qualitative evaluation.} In Fig. \ref{fig:visual_examples}, we show some qualitative tracking results of our method in comparison with eight top trackers, including KeepTrack \cite{Mayer2021LearningTC}, TrSiam \cite{Wang2021TransformerMT}, TrDiMP \cite{Wang2021TransformerMT}, SAOT \cite{Zhou2021SaliencyAssociatedOT}, KYS \cite{Bhat2020KnowYS}, PrDiMP50 \cite{danelljan2020probabilistic}, DiMP50 \cite{bhat2019learning}, DiMP18 \cite{bhat2019learning}. As can be seen, only our S-KeepTrack succeeds to maintain robustness in these four examples that subject to challenges including rotation, background clutter, illumination, partial occlusion, motion blur, low resolution, and scale variation. Specifically, all other trackers fail to keep track of the target in polo\_4, only S-KeepTrack and Trsiam succeed to track the target in basketball\_14, and only S-KeepTrack and KeepTrack succeed to track the target in 
golf\_4. We own the advantage of S-KeepTrack over other trackers, especially over KeepTrack, to the proposed method of combining low level and high level features for representation.

\begin{table*}[]
\centering
\caption{Illustration of the impact of different backbones on precision (PRC) and AUC on TSFMO. The better one is marked in bold. }
\label{tab:impact_of_backbone}\vspace{-0.1in}
\resizebox{3.6in}{0.23in}{
\begin{tabular}{lcccc} 
\toprule
          & DiMP                             & PrDiMP                           & KeepTrack                        & S-KeepTrack (ours)                 \\ 
\hline
ResNet18~ & ( 0.355, 0.197 )                   & ( 0.361, 0.204 )                   & ( 0.346, 0.185 )                   & ( 0.353,~ 0.190 )                    \\
ResNet50~ & (\textbf{0.366}, \textbf{0.206}) & (\textbf{0.390}, \textbf{0.225}) & (\textbf{0.425}, \textbf{0.247}) & (\textbf{0.437},~ \textbf{0.255})  \\
\bottomrule
\end{tabular}
}\vspace{-0.25in}
\end{table*}
\subsection{Ablation Study}
\textbf{Impact of backbone.} To study the impact of the backbone on performance of tracking small and fast moving object. We evaluate several state-of-the-art trackers with ResNet-18 and ResNet-50 as backbone separately, including DiMP \cite{bhat2019learning}, PrDiMP \cite{danelljan2020probabilistic}, KeepTrack \cite{Mayer2021LearningTC}, and our S-KeepTrack. Note that KeepTrack was implemented with only ResNet-50 as backbone originally. We adapt it and S-KeepTrack to support ResNet-18 as backbone for this ablation study. Table \ref{tab:impact_of_backbone} shows the PRC and AUC of these trackers on TSFMO in the form of (PRC, AUC). As can be seen, in each tracker the PRC and AUC are higher with ResNet-50 than with ResNet-18. Specifically, the (PRC, AUC) of DiMP, PrDiMP, KeepTrack, and S-KeepTrack increases by (1.1\%, 0.9\%), (2.9\%, 2.1\%), (7.9\%, 6.2\%), (8.4\%, 6.5\%) when the backbone is replaced from ResNet-18 to ResNet-50. This suggests that, although low-level features is helpful for tracking small and fast moving object as demonstrated by our method, deeper backbones are crucial to learn representation that extract abstract and essential information for this tracking task.

%


\begin{table*}[b]
\centering
\caption{Illustration of the impact of the importance coefficient of the low-level and high-level features on precision (PRC) and AUC on TSFMO. \textcolor{red}{Red}, \textcolor{blue}{blue} and \textcolor{green}{green} indicate the first, second and third place. }
\label{tab:impact_of_omega}\vspace{-0.1in}
\resizebox{4.0in}{0.22in}{
\begin{tabular}{cccccccccccc} 
\toprule
\textbf{$\omega$}       & 1.0   & 0.9   & 0.8   & 0.7   & 0.6   & 0.5   & 0.4                                        & 0.3                                        & 0.2                             & 0.1                              & 0.0                                           \\
\hline\hline
\textbf{PRC} & 0.424 & 0.417 & 0.415 & 0.424 & 0.421 & 0.427 & 0.428                                      & 0.428                                      & \textbf{\textcolor{red}{0.437}} & \textbf{\textcolor{blue}{0.433}} & \textcolor[rgb]{0,0.502,0}{\textbf{0.432}}  \\
\textbf{AUC}       & 0.247 & 0.241 & 0.240 & 0.246 & 0.243 & 0.247 & \textcolor[rgb]{0,0.502,0}{\textbf{0.248}} & \textcolor[rgb]{0,0.502,0}{\textbf{0.248}} & \textcolor{red}{\textbf{0.255}} & \textbf{\textcolor{blue}{0.251}} & \textbf{\textcolor{blue}{0.251}}            \\
\bottomrule
\end{tabular}
}	\vspace{-0.15in}
\end{table*}

\textbf{Impact of the importance coefficient of the low-level and high-level features.} To study the impact of the weight coefficient that balances the contributions of the similarity matrice $S$ and $\bar{S}$ estimated with low-level and high-level features, respectively, as formulated in Eq. (\ref{Eq:weight_coefficient}), we evaluate the proposed S-KeepTrack trained with different setting of the weight coefficient $\omega$ on TSFMO. The tried $\omega$ ranges from 0.0 to 1.0 with step size 0.1. Note that $\omega=0.0$ and $\omega=1.0$ mean using exclusively low-level and high-level features, respectively, where the latter is just the KeepTrack. Therefore, the larger the $\omega$ the more contributions of high-level features and the less of low-level one. The PRC and AUC of S-KeepTrack on TSFMO with respect to $\omega$ are shown on Table \ref{tab:impact_of_omega}. As can be seen, The highest PRC and AUC occur when $\omega$ is less than 
or equal to 0.4, suggesting, in a sense, that the more contributions of low-level features the higher the tracking performance. Specifically, when $\omega$=0.2, S-KeepTrack achieves the best PRC and AUC, i.e., 0.437 and 0.255, which is used as the default setting of $\omega$ for S-KeepTrack. When $\omega$=1.0 S-KeepTrack reduces to KeepTrack, having PRC and AUC of 0.424 and 0.247, respectively. S-KeepTrack surpasses KeepTrack on PRC and AUC by 1.3\% and 0.8\%, respectively, owing to introduced low-level features. Remarkably, we can observe that using low-level features is more effective than high-level features, as when $\omega$=0.0, i.e., using low-level features exclusively, S-KeepTrack achieves PRC and AUC of 0.432 and 0.251, surpassing KeepTrack with gaps of 0.8\% and 0.4\% on PRC and AUC, respectively.


\section{Conclusion}
In this work, we explore a new tracking task, i.e., tracking small and fast moving objects. In particular, we propose the TSFMO, which is the first benchmark for small and fast moving object tracking, to our best knowledge. In addition, in order to understand the performance of existing trackers and to provide baseline for future comparison, we extensively evaluate 20 state-of-the-art tracking algorithms with in-depth analysis. Moreover, we propose a novel tracker, named 
S-KeepTrack, by combining low-level and high-level features to obtain a stronger tracker, which outperforms existing state-of-the-art tracking algorithms by a clear margin. Our experiments suggest that there is a big room for us to improve the performance of tracking small and fast moving objects. We believe that, the benchmark, evaluation and the baseline tracker will inspire and facilitate more future research and application on tracking small and fast moving objects.

%
%



\bibliographystyle{splncs}
\bibliography{egbib}

\end{document}